\pdfoutput=1

\documentclass[11pt]{article}
\usepackage{colortbl}
\usepackage{xcolor}
\usepackage{amsmath}
\usepackage{tcolorbox}
\usepackage{array}
\usepackage{graphicx}
\usepackage{booktabs}

\definecolor{darkgreen}{rgb}{0.0, 0.5, 0.0}

\usepackage[preprint]{acl}

\usepackage{times}
\usepackage{latexsym}
\usepackage{CJKutf8}
\usepackage[T1]{fontenc}

\usepackage[utf8]{inputenc}

\usepackage{microtype}

\usepackage{inconsolata}

\usepackage{graphicx}

\title{Cultural Adaptation of Menus: A Fine-Grained Approach}

\author{
  Zhonghe Zhang, Xiaoyu He, Vivek Iyer, Alexandra Birch\\
  University of Edinburgh \\
  \texttt{zhonghe.zhang@hotmail.com, claire.xiaoyu.he@gmail.com,} \\
  \texttt{\{vivek.iyer, a.birch\}@ed.ac.uk}
}

\begin{document}
\maketitle

\begin{abstract}


Machine Translation of Culture-Specific Items (CSIs) poses significant challenges. Recent work on CSI translation has shown some success using Large Language Models (LLMs) to adapt to different languages and cultures; however, a deeper analysis is needed to examine the benefits and pitfalls of each method. In this paper, we introduce the ChineseMenuCSI dataset, the largest for Chinese-English menu corpora, annotated with CSI vs Non-CSI labels and a fine-grained test set. We define three levels of CSI figurativeness for a more nuanced analysis and develop a novel methodology for automatic CSI identification, which outperforms GPT-based prompts in most categories. Importantly, we are the first to integrate human translation theories into LLM-driven translation processes, significantly improving translation accuracy, with COMET scores increasing by up to 7 points. The code and datasets will be released shortly.

\end{abstract}

\section{Introduction}

\begin{figure}
    \centering
    \includegraphics[width=1\columnwidth]{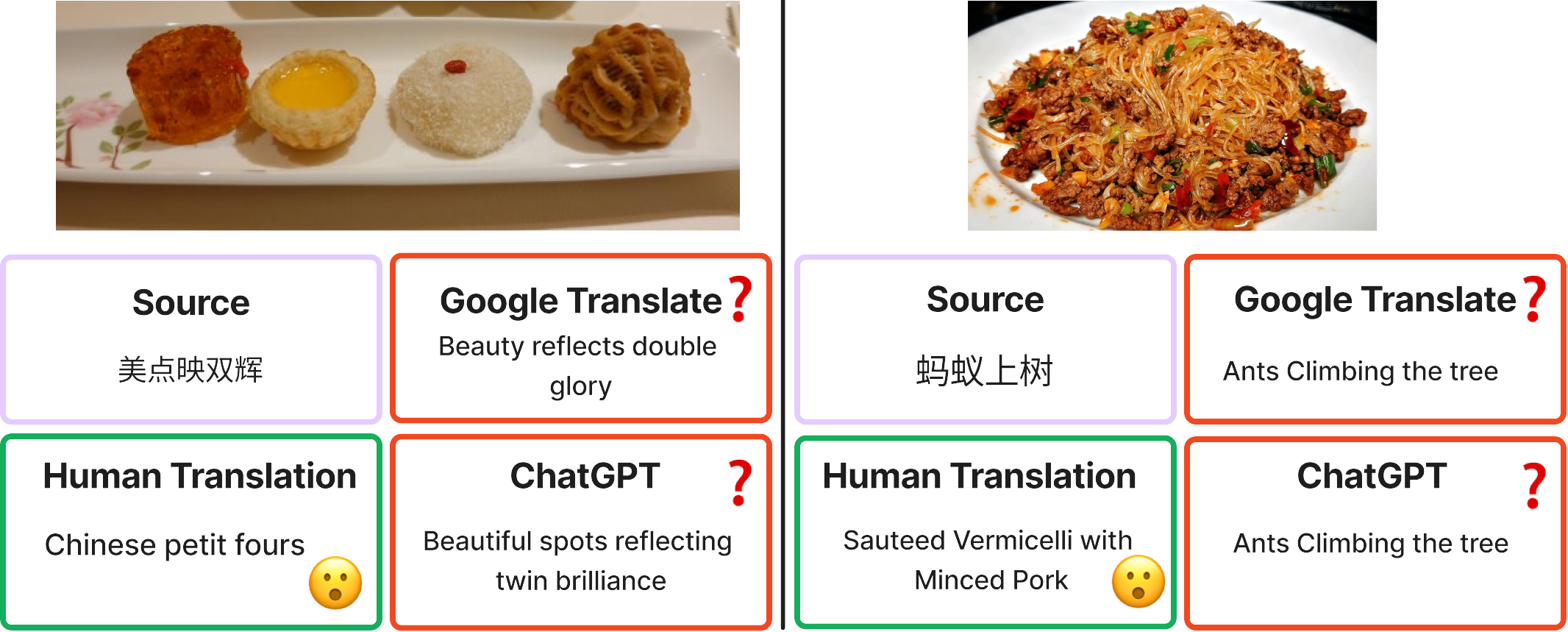}
    \caption{CSI translation errors by Google Translate and ChatGPT 3.5 in translating Chinese culinary terms}
    \label{fig:cultural_translation_error}
\end{figure}

Translating restaurant menus is a challenging, non-literal translation task. Unlike other texts, dish names are not merely lists of ingredients and culinary methods; they are short \citep{pellatt_and_liu_2010} and culturally rich expressions that require an understanding of cultural traditions \citep{amenador_wang_2022}, symbolism \citep{lam2018}, and local nuances. This complexity is compounded by LLMs and Neural Machine Translation (NMT) systems that often lack the cultural awareness necessary to accurately understand these nuances \citep{liu2023multilingual, naous2023having, tao2024cultural}. This results in mistranslations that can confuse and mislead the target audience \citep{garcea2023translate, 10.1145/3406324.3410721}, such as in Figure \ref{fig:cultural_translation_error}.

A key challenge in menu translation lies in the handling of Culture-Specific Items (CSIs), defined as ``concepts that are specific to a particular language or group” \citep{aixela1996culture}. For example, the literal translation of the Chinese dish \begin{CJK}{UTF8}{gbsn}蚂蚁上树\end{CJK} is ``Ants Climbing a Tree'' -- but this is actually a figurative Chinese expression that should be translated in English as ``Sauteed Vermicelli with Minced Pork''. The Chinese name creatively expresses the idea that pork resembles ants, while vermicelli represents tree branches. Existing machine translation systems, trained on plain, sentence-level translations, fail to capture these cultural subtleties and generate literal translations (Figure \ref{fig:cultural_translation_error}). 

However, there has been little work in NLP exploring CSI translations in-depth, particularly focusing on how the translation outputs generated by neural models should be improved. There has been foundational work on improving translations of CSIs by LLMs through enhanced prompting strategies \cite{yao_jiang_yang_hu}. Simultaneously, there has also been work on adapting CSIs  \cite{peskov-etal-2021-adapting-entities, cao_2024, singh2024translating}, but their focus has been on adapting culture-specific named entities. In this work, we seek to go beyond entities, and approach the translation of figurative language imbued with cultural nuance, as exemplified in Figure 1 -- which is quite underexplored.


In linguistics, cultural translation theories have been developed and widely adopted by human theorists and translators over decades. We aim to improve MT of CSIs by bringing the wisdom of Translation Theory research to modern NLP models like LLMs. Our approach improves the identification and translation of figurative and culturally nuanced CSIs. Unlike previous CSI identification methods, our method does not depend on parallel corpora or extensive knowledge graphs \citep{yao_jiang_yang_hu, han-etal-2023-bridging} -- but at the same time, we also show how recipes can be \emph{optionally} leveraged as a source of external knowledge to enhance performance even further. We also propose a novel CSI taxonomy for Chinese-English, that allows for a detailed analysis of figurative and culturally nuanced language and the translation challenges therein.  We evaluate our proposed methods using a large dataset of Chinese dish names, ensuring robust and reliable results.

Our key contributions are as follows:

\begin{enumerate}
    \item We introduce \textit{ChineseMenuCSI}, a fine-grained dataset of 4,275 bilingual Chinese-English restaurant menu entries from UK Chinese restaurants. The dataset is categorised into CSI and Non-CSI entities, with 480 entries further bifurcated into specific CSI categories, enabling an in-depth analysis of CSI translation efficacy of LLMs and NMT systems.

    \item We propose novel techniques for identifying CSIs, grounded in human translation theory. These techniques match or outperform current GPT-based prompts -- all without needing external knowledge graphs or parallel corpora.

    \item Lastly, we show how external knowledge, in the form of recipes, can add to the benefits pf translation strategies and enhance CSI translation performance further - achieving significant improvements in COMET scores, with gains of +3 to +7 points across CSI categories. 

\end{enumerate}

The code and datasets will be released shortly.

\section{Related Work}
\subsection{CSIs in Translation Studies}

\citet{aixela1996culture} was among the first scholars to introduce the term “culture-specific items” (CSIs) to refer to elements in texts that are unique and significant in a specific culture. CSIs may include objects, classification systems, or measurement tools common in the source culture but foreign to the target culture. Additionally, CSIs can encompass transcriptions of opinions/habits specific to a culture, which are often reflected in the language structure, style, and content. 

The concept of culture is closely related to understanding and translating CSIs, as \citet{aixela1996culture} highlighted. In the 1960s, \citet{nida_1963} introduced the concepts of formal and dynamic equivalence in translation to distinguish between structurally accurate and fluency-focused approaches to translation. These concepts have laid the foundation for subsequent translation theories, including those related to cultural translation. Expanding on this, \citet{newmark1988textbook} proposed a set of robust strategies for translating cultural elements, which have been particularly influential in translating Chinese culinary CSIs, as noted by \citet{amenador_wang_2022}. 

According to \citet{newmark1988textbook}, adaptation uses a recognized equivalent between two cultures. This strategy has been explored by \citet{pellatt_and_liu_2010} on Chinese menu translation. \citet{newmark1988textbook} proposed three equivalent strategies for translation: cultural, functional, and descriptive -- which we introduce later to improve LLM translation performance in §\ref{sec:translation-studies-prompting}.

Neutralisation is another translation strategy related to CSI translation. As proposed by \citet{inbook}, on the continuum between foreignisation (focusing on source culture) and domestication (focusing on target culture), there are intermediary approaches, including neutrality and neutralisation. For culture-specific text, neutralisation involves paraphrasing to convey the meaning of a CSI. After analysing the translations of Chinese dish names into English, \citet{amenador_wang_2022} found that neutralisation, by substituting the source text element with a more or less detailed explanation of its meaning, is the most commonly used translation strategy by human translators for translating Chinese dish names into English.

In this paper, we use these conventional translation strategies employed by human translators as instructions in zero-shot prompts, to enhance CSI translation quality of LLMs (in §\ref{sec:translation-studies-prompting}).

\subsection{Culture-Aware NMT}
Despite the early successes of NMT \cite{bahdanau2014neural, 10.5555/2969033.2969173}, translation of culture-specific texts has remained a daunting task. In addition to the challenge of translating rarer words and adapting to under-resourced domains \citep{koehn-knowles-2017-six}, CSIs are deeply intertwined with cultures \cite{hershcovich-etal-2022-challenges, liebling-etal-2022-opportunities, yao_jiang_yang_hu} -- something even the most capable neural models of today fail to grasp, particularly for non-Western cultures \citep{masoud2023cultural, alkhamissi2024investigating, nayak2024benchmarking}.

While there have been related works on domain-specific translation, including terminology translation \citep{dinu_etal_2019_training}, disambiguation \citep{iyer-etal-2023-code, iyer-etal-2023-towards} and named entity translation~\citep{hu_hayashi_cho_neubig_2022}, CSIs often lack direct equivalents in other languages, making translation complex and hard to understand cross-culturally \cite{yao_jiang_yang_hu}. 

Our approach uniquely combines translation studies with modern NLP techniques to identify and translate CSIs more effectively, resulting in more culturally sensitive and comprehensible translations.

\subsection{Cultural Awareness and Adaptation in Large Language Models}
In recent times, many works have shown that LLMs contain significant cultural biases against non-Western cultures \cite{cao2023assessing, liu2023multilingual, masoud2023cultural, naous2023having, tao2024cultural}. In response, there has been a growing focus on improving cultural awareness in LLMs through prompt-engineering techniques \cite{wang2023not, tao2024cultural} and fine-tuning on culture-specific data \cite{chan2023harmonizing, li2024culturellm, li2024culturepark}. Various tasks have been used to assess LLMs' cultural awareness, including tasks like culturally aware inference \cite{huang_yang_2023, yao_jiang_yang_hu} and common sense reasoning on specific languages \cite{koto2024arabicmmlu, koto2024indoculture}. 

Previous works on cultural awareness have primarily focused on understanding cultural norms in different languages rather than accurately translating culture-specific text. While well-explored in translation studies, cultural adaptation is rather understudied in NLP. Initial efforts in this direction have included adaptation of recipes \cite{cao_2024} and localisation of named entities through adaptation \cite{peskov-etal-2021-adapting-entities} or explicitation \cite{kementchedjhieva2020apposcorpus, garcea2023translate, han-etal-2023-bridging}. Most similar to our work is that of \citet{yao_jiang_yang_hu}, who also released a CSI dataset covering 6 languages, on which they benchmark LLMs and NMT systems. 
    
In contrast, our goal is to conduct a more fine-grained evaluation, given multiple CSI types in any given language. So, we leverage translation studies to create a dataset that classifies Chinese-English dishes into fine-grained categories, which we use for downstream evaluation, analysis and a detailed ablation of our proposed techniques. While we focus on the Chinese-English pair and culinary domain in this work, our framework and proposed techniques are agnostic of language/domain, and are designed to be easily scalable.

\section{ChineseMenuCSI Dataset}
We introduce a new bilingual Chinese-English Restaurant Menu (ChineseMenuCSI) dataset consisting of 4,275 human-verified dish entries collected from restaurants in UK.

\subsection{Data Collection}

We develop a Selenium-based web crawler\footnote{Selenium: \url{https://www.selenium.dev/}} to gather localized Chinese menu translations from restaurant websites in the UK. The selection criteria included restaurants with ratings above 3 out of 5 and an average meal price of over £20, ensuring that the menus would likely be high quality and not generated using commercial Machine Translation systems like Google Translate. These restaurants were sourced from TripAdvisor\footnote{TripAdvisor: \url{https://www.tripadvisor.com}}. 

We develop a heuristic parser to extract dish information from the bilingual menu images by detecting price tags and segmenting the raw text into aligned content. To achieve this, we utilise Google Cloud Vision OCR\footnote{Google Cloud Vision OCR: \url{https://cloud.google.com/vision}} to extract text and bounding boxes from the menu images. Price tags serve as unique indicators for each dish's content, as we observe that most menus included prices alongside their respective dishes. These price tags are identified using regular expressions, such as "dd.dd" or "£dd.dd".

Given that the position of price tags relative to dish names can vary across menus, we calculate alignment scores based on the cosine similarity and the gap distance between the potentially aligned Chinese and English text and select the alignment with the highest score from all possible combinations. Each entry undergoes manual review to ensure accuracy and errors are corrected before subsequent steps.

\subsection{CSI Taxonomy}
Translating Chinese menu items into English presents unique challenges because the dish names contain non-descriptive, picturesque elements \citep{pellatt_and_liu_2010}. Our initial data inspection revealed that CSIs within these dish names contribute varying degrees of complexity to the translation process, and carry differing levels of figurativeness brought by cultural and linguistic nuances. Inspired by translation theory literature that tends to categorise Chinese dishes into concrete and abstract categories \cite{lam2018}, we develop an approach to categorise the Chinese dish names in our dataset into three groups based on the degree of figurativeness in each CSI.

\paragraph{Category 1: Concrete CSIs (With a Low-level/No Figurative Meaning)}

\paragraph{Definition:} The CSIs in this category have a minimal figurative meaning, often referring to tangible attributes like ingredients, colour, taste, container, processing method, and dish appearance. Readers can easily understand these dish names as the information is either shared between the source and target cultures or has widely used translations in the target culture.

\paragraph{Example:} An example from the corpus is \textit{\begin{CJK}{UTF8}{gbsn}“咕噜猪肉”\end{CJK}} \textit{(sweet and sour pork)}. The first two Chinese characters \begin{CJK}{UTF8}{gbsn}“咕噜”\end{CJK} denote the Guangdong-style "sweet and sour" method, a culinary translation widely recognised outside the Chinese culture. The last two characters \begin{CJK}{UTF8}{gbsn}“猪肉”\end{CJK} mean "pork", a culturally universal ingredient.

\paragraph{Category 2: Creative CSIs (With Some Figurative Meaning)}

\paragraph{Definition:} This category features dish names that blend concrete lexical terms with figurative meanings, creating inventive expressions that extend beyond literal definitions. Understanding these dishes necessitates integration of creative flair with concrete information, presenting challenges.


\paragraph{Example:} \textit{\begin{CJK}{UTF8}{gbsn}“水煮鱼”\end{CJK}} (\textit{Poached fish fillet with chilli oil and herb} or \textit{Sichuan-style boiled fish}) originates from Sichuan, China. While the literal translation of the characters is "water-boiled fish", "water-boil" carries a creative description, representing the cooking state. This dish involves a Sichuan cooking style that uses hot chilli oil and Chinese herbs. "Poached fish fillet with chilli oil and herb" effectively describes the ingredients and cooking method, while "Sichuan-style boiled fish" adds cultural context by highlighting the dish's regional origin. Both are valid translations but different strategies are used.

\paragraph{Category 3: Abstract CSIs (With a High-level of Figurative Meaning)}

\paragraph{Definition:} This category encompasses dish names that exist beyond the realm of literal translation, and require in-depth cultural knowledge to understand. Crafted from metaphors, idioms, allegories, and other figurative language, these names disconnect from straightforward translations to engage in storytelling, aiming to convey broader narratives, evoke emotions, or reflect cultural heritage.

\paragraph{Example:} \textit{\begin{CJK}{UTF8}{gbsn}"佛跳墙”\end{CJK}} \textit{(Buddha Jumps Over the Wall }or\textit{ Steamed Abalone with Fish Maw in Chicken Broth)} metaphorically describes a dish so enticing that even a vegetarian and divine figure like Buddha would leap over a wall to taste it. Popular translations include "Buddha jumps over the wall" as a direct translation, and "Steamed Abalone with Fish Maw in Chicken Broth" includes ingredients and cooking methods, reflecting the dish's cultural and culinary nuances.

\subsection{Data Annotation}
\label{sec:dataannotation}

To annotate our data, we first seek to classify the data into CSI and non-CSI entities, and if it is a CSI, we want to categorise it into one of the above-listed groups. Given the dataset has as many as 4.3K CSIs, we approach the annotation process in two stages: a) in Stage 1, we conduct a broad, albeit rough, annotation of the entire dataset by two annotators, and b) we uniformly sample from the annotations of Stage 1 to ensure a fair distribution across categories, and conduct a more focused and rigorous annotation process using five translators -- to create our fine-grained test set. We describe these stages in more detail below: 

\paragraph{Stage 1 (Broad) Annotation:} Two annotators, who are postgraduate students and professional Chinese-English translators, reviewed and labelled all 4,275 entries. Both are native Chinese speakers proficient in English, ensuring high linguistic and cultural expertise. Firstly, we classify the entries into CSIs and non-CSIs. We use Cohen's kappa \citep{cohen1960coefficient} to measure agreement between the annotators and obtained a high score of 0.91 - likely because the classification of CSI and non-CSI is a mostly unambiguous task. For the 187 entries without consensus, we invited a third annotator to label these and assigned the final label using a majority vote. 

For entries with CSI, the annotators further categorised the dish into one of the three CSI categories. The annotation results are reported in Table \ref{table:first_csi_annoatation}.

\begin{table}[h!]
\centering
\small
\begin{tabular}{cll}
\hline
\noalign{\vspace{0.5mm}}
\textbf{Label} & \textbf{Category} & \textbf{Count} \\ 
\noalign{\vspace{0.5mm}}
\hline
\noalign{\vspace{0.5mm}}
0 & Non-CSIs & 2003 \\ 
1 & Concrete CSIs  & 1658 \\ 
2 & Creative CSIs & 494 \\ 
3 & Abstract CSIs & 120 \\ 
\hline
\end{tabular}
\caption{Distribution of menu items across CSI taxonomy in the ChineseMenuCSI dataset}
\label{table:first_csi_annoatation}
\end{table}

\paragraph{Stage 2 (Focused) Annotation: } In Table \ref{table:first_csi_annoatation}, we note that Category 3 is the smallest, with only 120 items. To evenly balance our test set, we randomly sample 120 items from each category: 0 (Non-CSIs), 1 (Concrete CSIs), 2 (Creative CSIs), and 3 (Abstract CSIs), totalling 480 items. This subset was annotated by a larger and more diverse group of five annotators, who, like the first-stage annotators, included professional Chinese-English translators and postgraduate students -- all native Chinese speakers proficient in English.


For inter-annotator agreement, we use Fleiss' kappa \citep{fleiss1971measuring} across two levels: CSI fine-grained categorisation and span-level CSI identification; the results are summarised in Table \ref{table:csi_annotation_agreement_score}.

\begin{table}[h!]
\centering
\small
\begin{tabular}{lcc}
\hline
\noalign{\vspace{0.5mm}}
\textbf{Annotation} & \textbf{Kappa} & \textbf{Interpretation} \\
\noalign{\vspace{0.5mm}}
\hline
\noalign{\vspace{0.5mm}}
CSI vs. Non-CSI & 0.91 & High \\
CSI Category & 0.63 & Substantial \\
CSI Identification & 0.70 & Substantial \\
\hline
\end{tabular}
\caption{Inter-annotator agreement scores for different annotation tasks}
\label{table:csi_annotation_agreement_score}
\end{table}

For the fine-grained CSI categorisation, we exclude items that do not attain majority consensus (at least 3 out of 5 annotators in agreement), resulting in a kappa score of 0.63. This score falls within the range of substantial agreement (0.6-0.8) but is lower than the CSI vs. Non-CSI score due to the subjective nature of the fine-grained categorisation. This level of agreement is comparable to ranges reported in related work \citep{huang_yang_2023, soderstrom2021developing}. Lastly, for CSI identification at the span-level, i.e. within a given dish name, the kappa score is 0.70 - which indicates substantial agreement as well.

\section{CSI Automatic Identification}

\label{sec:methodology_csi_identification}

To accurately translate CSIs, it is essential to first identify which parts of the text comprise CSIs. Previous studies have approached this challenge in different ways. \citet{han-etal-2023-bridging} focus on implicit detection and used a relative distance of terms in Wikidata, but it does not include all Chinese dish CSI. \citet{yao_jiang_yang_hu} rely on parallel corpora with entity-linking to find CSIs; however, the approach is infeasible for online MT, where we need to identify CSIs beforehand to produce translations from the monolingual source text.


Inspired by these methods, we propose a method called \textbf{Combined CSI Identification}, that uses a combination of three checking criteria for CSI identification, and classifies CSI if at least two of the following three checks are met. The checks are Round-trip Translation (RTT), Cultural Uniqueness (CU), and Historical Significance (HS).

\subsection{Round-trip Translation (RTT)}

Since CSIs are defined as terms unique to a specific language or culture \citep{alvarez1996translation}, based on the assumption that they do not have corresponding translations in the target language, we propose using round-trip translation (RTT) as one of the identifying criteria.

\begin{enumerate}
\item \textbf{Initial Translation}: Translate the Chinese dish name to English using Google Translate.
\item \textbf{RTT Translation}: Translate the English version back to Chinese using DeepL Translate and split it into Jieba-segmented words\footnote{Jieba: \url{https://github.com/fxsjy/jieba} }. Using different translation systems for RTT proved most effective in identifying CSIs.
\item \textbf{Identification}: Subtract the segmented words in the RTT from those in the original text. The remaining words are potential CSIs. 

\[
\small\text{CSIs} = \text{Original Words} - \text{RTT Words}
\]

Using Jieba's cut-for-search module, which returns all words and phrases, a phrase is considered CSI only if all of its words are omitted in the RTT, and not otherwise.

\end{enumerate}

This method certainly has its limitations, for example it could also: a) return words that are not CSIs and are just difficult to translate, and b) miss CSIs that have literal translations. Still, we find in §\ref{sec:evaluation_csi_identification} that it performs strongly in identifying CSIs in most cases.

\subsection{Cultural Uniqueness (CU)}

According to \citet{newmark1988textbook}, ``unfindable'' words are often less frequently seen within a language. Words with cultural and historical references can be deeply embedded in a specific culture or history, making them rare or unfamiliar to outsiders.

We use Jieba to segment words in the ChineseMenuCSI dataset, then measure each word's frequency and calculate its inverse frequency. A cut-off at the 95th percentile of these inverse frequencies is set based on a manual review of 100 words. Words above this cut-off are marked as potential CSI. Words not previously seen are given an inverse frequency of 1, indicating they are potential CSI. No smoothing techniques are applied, as inverse frequency is used against a fixed threshold rather than for probability calculations.

\subsection{Historical Significance (HS)}

Chinese food names and CSIs often contain historical narratives such as historical events, figures and periods \cite{lam2018, amenador_wang_2022}. To identify these, we use the Wikipedia API to search for individual words or entire dish names. If a word's Wikipedia page includes a "History" section, it is considered a potential CSI. The words appearing 30 times or more, such as "chicken" or "sauce," are excluded as generic terms.


\section{CSI Translation}

Having identified CSIs, we propose some prompting strategies to improve CSI translation performance. Our strategies fall into two categories: Recipe-based Translation and Translation Studies-inspired Prompting. 

\subsection{Recipe-based Translation}

\begin{figure*}[t]
    \centering
    \includegraphics[width=\linewidth]{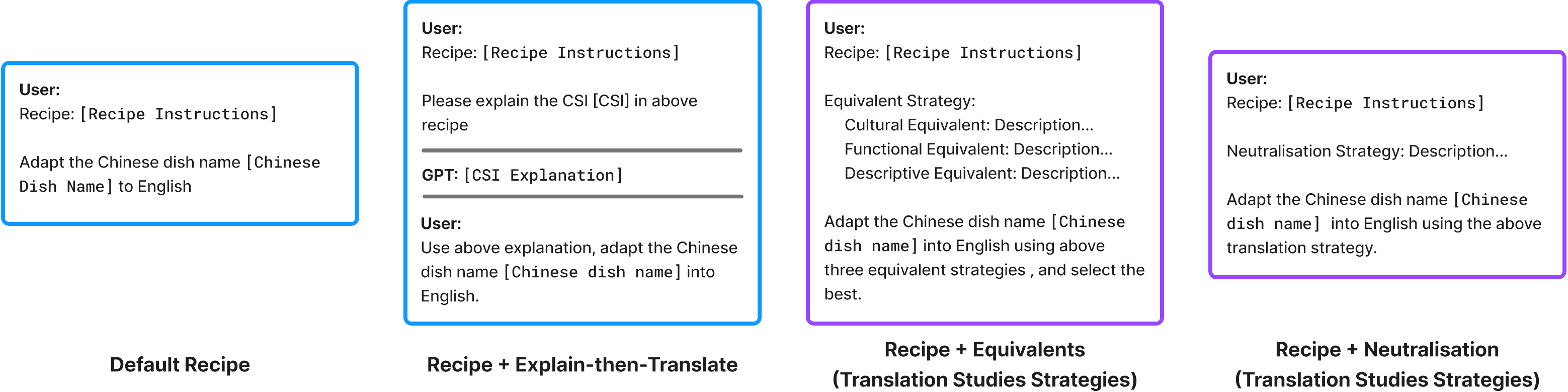}
    \caption{Four adaptation prompt strategies}
    \label{fig:adapation_prompt_stragtegies}
\end{figure*}

We explore using recipe information to improve the translation of CSI dish names. By incorporating the most relevant recipe as external knowledge, we experimented with two zero-shot prompt strategies: \textbf{Default Recipe} prompting and \textbf{Recipe + Explain-then-Translation} prompting (Figure \ref{fig:adapation_prompt_stragtegies}).

\paragraph{Recipe Retrieval Pipeline}
\label{sec:prompt}

Given over 50\% of the CSI dish entries in our test set lack detailed publicly available descriptions from sources like Wikipedia, we use the Xiachufang recipe database \cite{liu2022counterfactual} -- which contains approximately 1.4 million Chinese monolingual recipes -- to retrieve recipes and enhance translation accuracy. 

In proposing this approach, we take inspiration from research in Translation Studies that has highlighted the importance of cooking methods and ingredients in translating Chinese dish names \citep{amenador_wang_2022}.

Our retrieval pipeline involves two key stages:
\begin{enumerate}
    \item \textbf{Query and document Construction:} The query is the full dish name, with the CSI span from previous annotations. We concatenated each recipe name and instructions into a single recipe document.
    \item \textbf{Filtering and Ranking Recipes:} To filter and rank the recipes, we employ the BM25 algorithm \cite{bm25}, which assigns a score to each word in the recipe document based on its term frequency and inverse document frequency. For each word in the recipe document, the score is enhanced by applying a weighting factor when the word matches either the dish name (weight = 5) or the CSI span (weight = 3), with an additional multiplier of 3 applied to words within the dish name to prioritize their importance. If there is no exact match for the dish name, the process shifts focus to matching with the CSI span. Additionally, we apply a length penalty to the final score, adjusting it based on the difference between the recipe’s length and the average length of all recipes. We select the top-ranked recipe as the final output.
    
\end{enumerate}

\paragraph{Prompt Strategy: CSI Recipe}
\label{sec:csi_recipe}

In this prompting strategy, we use the most relevant recipe returned from the aforementioned search pipeline to aid CSI translation. We provide the name and cooking instructions of the closest-matching recipe while noting that it might not correspond exactly to the given dish but is beneficial as external knowledge since it contains the CSIs to translate.

\paragraph{Prompt Strategy: CSI Explain-then-Translate}

Inspired by Chain of Thought (CoT) prompting \citep{wei_wang_schuurmans_bosma_ichter_xia_chi_quoc_le_zhou_2023} and Self-Explanation \citep{yao_jiang_yang_hu}, we formulate another prompting strategy that first asks the LLMs to explain the meaning of the CSIs described in the recipe and then generate the translation for the dish. The motivation is to help the LLMs conduct advanced reasoning on the recipe instructions, such as interpreting dish names with CSIs not explicitly defined in the recipe. For example, a recipe might instruct to ``cut it first, then stir fry" or note that ``it can be very spicy" without explaining the CSI. The LLM's task is to infer the meaning of the CSIs based on the recipe's instructions. The prompt is shown in Figure \ref{fig:adapation_prompt_stragtegies}.

\subsection{Translation Studies-inspired Prompting}
\label{sec:translation-studies-prompting}
Unlike the conventional prompt engineering used in related work, our second set of prompting strategies differs in that they incorporate \textit{human translation strategies}, inspired by the rich literature in Translation Studies, directly into the design of the prompt. We provide the prompt template for both of these strategies in Figure \ref{fig:adapation_prompt_stragtegies} and complete prompt in Appendix \ref{fig:prompt_template_equivalent} and \ref{fig:prompt_template_neutralisation}.

\paragraph{Prompt Strategy: Equivalents}
Using the recipe in §\ref{sec:csi_recipe} as external knowledge, we ask the LLMs to produce three translations, each based on a different translation strategy (i.e. cultural, functional and descriptive equivalents), and then task it to \textbf{select the best translation}. 

These translation strategies are inspired by \citet{newmark1988textbook}'s theories on equivalent translation. We define the equivalent strategies below, providing examples for the reader's understanding:

\begin{enumerate}
    \item \textbf{Cultural Equivalent:} Replacing a CSI in the source text with a term that is culturally relevant and functionally equivalent in the target culture. This strategy aims to evoke the same response in the target audience. (i.e. translating \begin{CJK}{UTF8}{gbsn}``粽子"\end{CJK} as ``tamale" in Spanish -- given both are traditional wrapped food items made with a starchy substance and fillings, albeit from different cultures.)
    \item \textbf{Functional Equivalent:} This strategy focuses on the function or purpose of the item (i.e., translating \begin{CJK}{UTF8}{gbsn}"粽子"\end{CJK} as ``rice dumpling" to convey the idea of a food made of rice)
    \item \textbf{Descriptive Equivalent:} Providing a detailed description or explanation of the CSI to convey its meaning and significance. This approach is useful when the CSI is essential for understanding the text but has no equivalent in the target language (i.e. translating \begin{CJK}{UTF8}{gbsn}"粽子"\end{CJK} as "a traditional Chinese sticky rice dumpling wrapped in bamboo leaves")
\end{enumerate}

\paragraph{Prompt Strategy: Neutralisation}

Another human translation strategy we used in the prompting experiments is neutralisation. Again, we provide recipe information as external knowledge as in the previous strategy and incorporate an explanation of the neutralisation strategy to guide the LLM translation of dish names. 

\textbf{Neutralisation:} Using culturally neutral language to describe or explain a cultural word, phrase, or rhetorical expression from the source text. It answers the question, ''What is this?" \citep{amenador_wang_2022} by adding information such as ingredients, culinary methods and key characteristics. Compared with the descriptive equivalent strategy, the neutralisation strategy we used for prompt design confines the information used in the translations to ingredients, culinary methods and key characteristics (i.e. translating \begin{CJK}{UTF8}{gbsn}"粽子"\end{CJK} as "sticky rice wrapped in bamboo leaves").

\section{Results and Analysis}
Our experiments consist of five parts: 1. Baseline evaluation of MT performance using three models on the ChineseMenuCSI dataset (§\ref{sec:evaluation_csi_vs_non_csi}); 2. Assessment of CSI span identification accuracy (§\ref{sec:evaluation_csi_identification}); 3. Exploration of main adaptation strategies (§\ref{sec:evaluation_csi_adaptation}); 4. Exploration of individual equivalents translation strategies for enhancing CSI translation (§\ref{sec:evauation_translation_theory}); 5. Human evaluation of translation quality on a subset of the dataset (§\ref{sec:evaluation_human}).

\subsection{Experimental Settings}

We evaluate the effectiveness of LLM translations for CSIs by comparing various prompting strategies across two SOTA LLMs — GPT-3.5 (gpt-3.5-turbo-0125\footnote{GPT-3.5:\url{https://platform.openai.com/docs/models/gpt-3-5-turbo}})  and the advanced GPT-4o (gpt-4o-2024-05-13\footnote{GPT-4o:\url{https://platform.openai.com/docs/models/gpt-4o}})  — against the robust commercial MT system, Google Translate. This approach allows us to assess the strengths of LLM prompting versus a widely used commercial MT.


    

\subsection{Evaluation of CSIs vs Non-CSIs}
\label{sec:evaluation_csi_vs_non_csi}
We use the Stage 1 annotated version of the ChineseMenuCSI dataset (§\ref{sec:dataannotation}). This version has a high inter-annotator agreement of 0.91 for the CSI vs non-CSI classification task, with conflicts further resolved using a third annotator. We compare the MT performance using the COMET version wmt22-comet-da \citep{rei_etal_2022_comet}.


\begin{table}[htbp]
\centering
\small
\begin{tabular}{lcc}
\hline
\noalign{\vspace{0.5mm}}
\textbf{Method} & \textbf{Non-CSIs} & \textbf{CSIs} \\
\noalign{\vspace{0.5mm}}
\hline
\noalign{\vspace{0.5mm}}
Google Translate & \textbf{74.08} & 64.48 \\
GPT-3.5 & 72.88 & 64.34 \\
GPT-4o & 73.67 & \textbf{65.97} \\
\hline
\end{tabular}
\caption{Comparison of COMET scores across different systems for CSI and Non-CSI menu items}
\label{tab
}
\end{table}




While GPT-4o yields major improvements in the CSI translation performance, GPT-3.5 and GPT-4o show worse scores for non-CSIs than Google Translate. These results suggest that Google Translate is particularly strong at translating straightforward, culturally neutral content, likely due to its extensive and diverse training dataset which prioritizes general language accuracy over cultural nuances. 


Interestingly, despite Google Translate's general strength in multilingual tasks \cite{zhu2023multilingual}, GPT-4o shows better handling of CSIs, highlighting the benefits of pretraining at scale on diverse corpora from many cultures, as opposed to NMT systems that are typically trained on narrow-domain sentence-level parallel corpora.

\subsection{Evaluation of CSI Span Identification }
\label{sec:evaluation_csi_identification}

We further assess the capability of different methods to \textit{pinpoint specific CSI spans} -- a task we call \textbf{CSI Span Identification} -- within dish names. This fine-grained analysis is crucial for understanding the elements that contribute to cultural specificity and translation complexity.

\begin{table}[t]
\small 
\centering

\begin{tabular}{llccc}

\hline
\noalign{\vspace{0.5mm}}
\textbf{} & \textbf{Method} & \textbf{Precision} & \textbf{Recall} & \textbf{F1 Score} \\
\noalign{\vspace{0.5mm}}
\hline
\noalign{\vspace{0.5mm}}
\multicolumn{5}{c}{\textbf{CSI-1: Concrete CSIs}} \\
\noalign{\vspace{0.5mm}}
\hline
\noalign{\vspace{0.5mm}}
Ours & Combined & 64.9\ & 34.0\ & 44.7\ \\
 & RTT & 58.8\ & 28.4\ & 38.3\ \\
  & CU & 38.3\ & 65.3\ & 48.2\ \\
 & HS & \textbf{86.3}\ & 31.2\ & 45.8\ \\
\hline
\noalign{\vspace{0.5mm}}
GPT & 3.5 & 32.4\ & \textbf{80.4}\ & 46.2\ \\
 & 4o & 37.9\ & 67.1\ & \textbf{48.5}\ \\
\hline
\noalign{\vspace{0.5mm}}
\multicolumn{5}{c}{\textbf{CSI-2: Creatives CSIs}} \\
\noalign{\vspace{0.5mm}}
\hline
\noalign{\vspace{0.5mm}}
Ours & Combined & 66.1\ & 53.1\ & 58.9\ \\
 & RTT & 63.4\ & 60.1\ & \textbf{61.7}\ \\
 & CU & 35.4\ & 70.4\ & 47.1\ \\
 & HS & \textbf{68.6}\ & 16.4\ & 26.5\ \\
\hline
\noalign{\vspace{0.5mm}}
GPT & 3.5 & 34.1\ & \textbf{82.0}\ & 48.2\ \\
 & 4o & 40.6\ & 73.9\ & 52.4\ \\
\hline
\noalign{\vspace{0.5mm}}
\multicolumn{5}{c}{\textbf{CSI-3: Abstract CSIs}} \\
\noalign{\vspace{0.5mm}}
\hline
\noalign{\vspace{0.5mm}}
Ours & Combined & 81.4\ & 68.6\ & 74.4\ \\
 & RTT & \textbf{81.7}\ & 73.6\ & \textbf{77.4}\ \\
 & CU & 43.9\ & 88.4\ & 58.6\ \\
 & HS & 80.0\ & 9.9\ & 17.6\ \\
\hline
\noalign{\vspace{0.5mm}}
GPT & 3.5 & 50.4\ & \textbf{94.3}\ & 65.7\ \\
 & 4o & 59.1\ & 78.9\ & 67.6\ \\
\hline
\end{tabular}
\caption{Evaluation of CSI span identification accuracy by CSI category: precision, recall, and F1 scores}
\label{tab:evaluation_csi_span_identification}
\end{table}

\begin{table*}[h]
\centering
\small
\begin{tabular}{lcccccccc}
\hline
\noalign{\vspace{0.5mm}}
& \multicolumn{4}{c}{\textbf{GPT-3.5}} & \multicolumn{4}{c}{\textbf{GPT-4o}} \\
\cmidrule(r){2-5} \cmidrule(r){6-9}
& \textbf{CSI-1} & \textbf{CSI-2} & \textbf{CSI-3} & \textbf{Overall} & \textbf{CSI-1} & \textbf{CSI-2} & \textbf{CSI-3} & \textbf{Overall} \\
\noalign{\vspace{0.5mm}}
\hline
\noalign{\vspace{2mm}}
Original  & 62.68 & 55.38 & 43.92 & 53.33 & 62.68 & 55.38 & 43.92 & 53.33 \\
\hline
\noalign{\vspace{1mm}}
\multicolumn{9}{l}{\textit{Recipe-based Translation}} \\

Recipe & \cellcolor{yellow!25}+0.16 & \cellcolor{red!10}-0.90 & \cellcolor{green!25}+3.44 & \cellcolor{yellow!25}+0.50 & \cellcolor{red!10}-0.08& \cellcolor{red!10}-3.02 & \cellcolor{green!25}+3.49 & \cellcolor{green!25}+1.93\\
Recipe + EtT & \cellcolor{yellow!25}+1.13 & \cellcolor{red!10}-1.33  & \cellcolor{green!50}+4.92 & \cellcolor{yellow!25}+1.04 & \cellcolor{green!25}+1.10 & \cellcolor{green!25}+1.61 & \cellcolor{green!50}+4.87 & \cellcolor{green!25}+2.16\\

\hline
\noalign{\vspace{1mm}}
\multicolumn{9}{l}{\textit{Translation Studies Prompting}} \\
Neutralisation & \cellcolor{yellow!25}+0.74 & \cellcolor{yellow!25}+1.15 & \cellcolor{green!50}+3.62 & \cellcolor{green!25}+1.56 & \cellcolor{yellow!25}+0.46 & \cellcolor{green!50}+\textbf{4.84} & \cellcolor{green!50}+4.29 &  \cellcolor{green!25}+3.02 \\
Equivalents & \cellcolor{yellow!25}+\textbf{1.44} & \cellcolor{green!25}+3.24 & \cellcolor{green!25}+2.52 & \cellcolor{green!25}+2.38 & \cellcolor{green!25}+\textbf{2.34} & 
\cellcolor{green!50}+3.89 &
\cellcolor{yellow!25}+0.94 &  \cellcolor{green!25}+2.62 \\

\hline
\noalign{\vspace{1mm}}
\multicolumn{9}{l}{\textit{Recipe + Translation Studies Prompting}} \\
Recipe + Neutralisation  & \cellcolor{red!10}-1.29 & \cellcolor{yellow!25}+1.15 & \cellcolor{green!100}\textbf{+7.72} & \cellcolor{green!25}+1.71 & \cellcolor{yellow!25}+0.09 & \cellcolor{green!25}+3.47 & \cellcolor{green!50}+4.25 & \cellcolor{green!25}+2.34 \\
Recipe + Equivalents & \cellcolor{yellow!25}+0.95 & \cellcolor{green!25}+\textbf{3.85} & \cellcolor{green!25}+3.01 & \cellcolor{green!25}+\textbf{2.54} & \cellcolor{green!25}+1.80 & \cellcolor{green!25}+3.24 & \cellcolor{green!100}\textbf{+7.87} & \cellcolor{green!50}\textbf{+3.74} \\

\hline
\end{tabular}

\caption{\small COMET score comparisons for GPT-3.5 and GPT-4o using various translation strategies across CSI categories. The overall score is calculated as the average of CSI-1, CSI-2, and CSI-3 scores for each method.}
\label{tab:evaluation_recipe_adaptaion_main_methods}
\end{table*}

\begin{table*}[t]
\centering
\small 
\begin{tabular}{lcccccccc}
\hline
\noalign{\vspace{0.5mm}}
& \multicolumn{4}{c}{\textbf{GPT-3.5}} & \multicolumn{4}{c}{\textbf{GPT-4o}} \\
\cmidrule(r){2-5} \cmidrule(r){6-9}
& \textbf{CSI-1} & \textbf{CSI-2} & \textbf{CSI-3} & \textbf{Overall} & \textbf{CSI-1} & \textbf{CSI-2} & \textbf{CSI-3} & \textbf{Overall} \\
\noalign{\vspace{0.5mm}}
\hline
\noalign{\vspace{2mm}}
Original  & 62.68 & 55.38 & 43.92 & 53.33 & 63.43 & 54.50 & 47.50 & 55.14 \\
\hline
\noalign{\vspace{1mm}}
\multicolumn{9}{l}{\textit{Equivalents Strategy Prompting}} \\
Cultural & \cellcolor{red!10}-0.06 & \cellcolor{red!10}-0.86 & \cellcolor{yellow!25}+0.99 & \cellcolor{yellow!25}+0.02 & \cellcolor{yellow!25}\textbf{+0.91} & \cellcolor{yellow!25}+0.90 & \cellcolor{red!10}-2.77 & \cellcolor{red!10}-0.32 \\
Descriptive & \cellcolor{red!10}-6.73 & \cellcolor{red!10}-1.90 & \cellcolor{yellow!25}+0.93 & \cellcolor{red!10}-2.57 & \cellcolor{red!10}-3.83 & \cellcolor{green!25}+2.62 & \cellcolor{green!25}+2.10 & \cellcolor{yellow!25}+0.96 \\
Functional & \cellcolor{yellow!25}+0.54 & \cellcolor{green!25}+2.69 & \cellcolor{yellow!25}+0.78 & \cellcolor{yellow!25}+1.34 & \cellcolor{yellow!25}+0.06 & \cellcolor{green!50}+3.47 & \cellcolor{green!25}+1.09 & \cellcolor{yellow!25}+1.54 \\
\hline
\noalign{\vspace{1mm}}
\multicolumn{9}{l}{\textit{Recipe + Equivalents Strategy Prompting}} \\
Recipe + Cultural & \cellcolor{red!10}-2.56 & \cellcolor{red!10}-0.89 & \cellcolor{red!10}-0.06 & \cellcolor{red!10}-1.83 & \cellcolor{yellow!25}+0.84 & \cellcolor{yellow!25}+1.73 & \cellcolor{yellow!25}+0.72 & \cellcolor{yellow!25}+1.10 \\
Recipe + Descriptive & \cellcolor{red!10}-8.69 & \cellcolor{red!10}-1.81 & \cellcolor{green!25}+2.29 & \cellcolor{red!10}-2.74 & \cellcolor{red!10}-4.74 & \cellcolor{green!50}\textbf{+5.27} & \cellcolor{green!50}+3.86 & \cellcolor{yellow!25}+1.46 \\
Recipe + Functional & \cellcolor{green!25}\textbf{+2.27} & \cellcolor{green!50}\textbf{+4.02} & \cellcolor{green!25}\textbf{+2.57} & \cellcolor{green!25}\textbf{+2.95} & \cellcolor{red!10}-0.96 & \cellcolor{green!25}+2.80 & \cellcolor{green!100}\textbf{+7.97} & \cellcolor{green!50}\textbf{+3.27} \\
\hline
\end{tabular}

\caption{\small Ablation study comparing COMET scores for GPT-3.5 and GPT-4o using different equivalent strategies across CSI categories. The overall score is calculated as the average of CSI-1, CSI-2, and CSI-3 scores for each method.}
\label{tab:equivalent_strategy_comparison}
\end{table*}

Table \ref{tab:evaluation_csi_span_identification} shows GPT 4o as the best performer in CSI-1 and RTT as the best in other categories. This is likely due to RTT's strength in identifying CSIs, which are often figurative and lack general interpretations. However, the combined metrics only improved RTT's performance in CSI-1, likely because of low recall in HS. Moreover, CU under-performs in CSI-2, as those CSIs typically involve figurative messages in creative combinations of frequent words, which cannot be captured by frequency.

The combined method does not outperform the individual highest method as it requires majority agreement, where a single correct check is insufficient. HS shows high precision but low recall, likely because CSIs often have historical backgrounds, though the inverse is not always true. 


\subsection{Evaluation of Main Adaptation Strategies}
\label{sec:evaluation_csi_adaptation}

This section examines whether external knowledge, such as recipes, can enhance MT of CSI-rich dish names. Four prompting strategies were tested: a) Default Recipe prompting, b) Recipe + Explain-then-Translate (EtT) prompting, c) Equivalents and d) Neutralisation. For the latter two, which are translation strategies, we try baselines with and without incorporation of recipes, to ablate the dependency on external knowledge. In Table \ref{tab:evaluation_recipe_adaptaion_main_methods}, we see that while all our proposed methods yield overall improvements in performance, translation strategy-based methods -- that do not involve any external knowledge -- yields the largest gains, of up to +4.8 COMET points! Moreover, when recipes are added to these strategies, the gap widens even further, with the maximum gain reaching \emph{as high as +7.87 COMET points} for our best performing \textit{Recipe + Equivalents} strategy. The second trend we note is that the largest gains come from the more complex CSI-2 and CSI-3 categories, indicating the efficacy of our translation theory-inspired methods for translating highly culturally nuanced text. Finally, while the more advanced model GPT-4o naturally yields the largest improvements, we note that we get pretty good results with the far cheaper GPT-3.5 model too, indicating that our methods could be used quite economically.


Revisiting the relatively lower improvements in CSI-1 by examining GPT-generated translations, we find that the LLMs sometimes focus on irrelevant details in the provided recipes. In CSI-1, which involves shorter CSI terms, finding an exact match for dish names is harder, forcing the inclusion of noise in the recipe. For instance, the term \begin{CJK}{UTF8}{gbsn}"咕噜”\end{CJK} \textit{(Sweet and sour)} applies to various dishes like pork, chicken, or fish, making it difficult to provide the correct ingredient. In contrast, CSI-2 and CSI-3 usually involve longer, more specific phrases like \begin{CJK}{UTF8}{gbsn}"蚂蚁上树”\end{CJK} \textit{(Fried vermicelli with pork)}, making it easier to find an exact recipe match, reduce noise, and majorly improve accuracy.

\begin{table*}[t]
\centering
\small
\begin{tabular}{lcccccccc}
\hline
\noalign{\vspace{0.5mm}}
& \multicolumn{4}{c}{\textbf{GPT-3.5}} & \multicolumn{4}{c}{\textbf{GPT-4o}} \\
\cmidrule(r){2-5} \cmidrule(r){6-9}
& \textbf{CSI-1} & \textbf{CSI-2} & \textbf{CSI-3} & \textbf{Overall} & \textbf{CSI-1} & \textbf{CSI-2} & \textbf{CSI-3} & \textbf{Overall} \\
\noalign{\vspace{0.5mm}}
\hline
\noalign{\vspace{2mm}}
Original & 6.33 & 3.88 & 3.18 & 4.47 & 6.22 & 4.23 & 3.65 & 4.70 \\
\hline
\noalign{\vspace{1mm}}
\multicolumn{9}{l}{\textit{Recipe-based Translation}} \\
Recipe & \cellcolor{red!10}-0.93 & \cellcolor{yellow!25}+0.67 & \cellcolor{green!25}+1.74 & \cellcolor{yellow!25}+0.49 & \cellcolor{red!10}-0.04 & \cellcolor{yellow!25}+0.80 & \cellcolor{green!50}+2.28 & \cellcolor{green!25}+1.01 \\
Recipe + Explain-then-Translate & \cellcolor{red!10}-0.03 & \cellcolor{yellow!25}+0.60 & \cellcolor{green!25}+1.35 & \cellcolor{yellow!25}+0.64 & \cellcolor{yellow!25}+0.43 & \cellcolor{yellow!25}+0.98 & \cellcolor{green!25}+1.68 & \cellcolor{green!25}+1.03 \\
\hline
\noalign{\vspace{1mm}}
\multicolumn{9}{l}{\textit{Recipe + Translation Studies Prompting}} \\
Recipe + Functional & \cellcolor{red!10}-1.15 & \cellcolor{green!25}+1.10 & \cellcolor{green!25}+1.71 & \cellcolor{yellow!25}+0.60 & \cellcolor{yellow!25}+0.75 & \cellcolor{green!25}+1.77 & \cellcolor{green!50}+2.14 & \cellcolor{green!25}+1.05 \\
Recipe + Neutralisation & \cellcolor{yellow!25}\textbf{+0.37} & \cellcolor{green!25}\textbf{+1.21} & \cellcolor{green!25}+1.03 & \cellcolor{yellow!25}+0.62 & \cellcolor{green!25}\textbf{+1.32} & \cellcolor{green!50}\textbf{+2.83} & \cellcolor{green!100}\textbf{+3.20} & \cellcolor{green!25}+\textbf{1.95}  \\
Recipe + Equivalent & \cellcolor{red!10}-0.85 & \cellcolor{yellow!25}+0.62 & \cellcolor{green!50}\textbf{+2.05} & \cellcolor{yellow!25}\textbf{+0.81} & \cellcolor{yellow!25}+0.71 & \cellcolor{yellow!25}+0.99 & \cellcolor{green!50}+2.38 & \cellcolor{green!25}+1.36 \\
\hline
\end{tabular}

\caption{\small Difference in human evaluation of translation quality compared to baseline for different models and strategies across CSI categories. The overall score is calculated as the average of CSI-1, CSI-2, and CSI-3 scores for each method.}
\label{tab:evaluation_recipe_adaptaion}
\end{table*}

\subsection{Evaluation of Individual Equivalent Strategies}
\label{sec:evauation_translation_theory}

We further perform an ablation analysis of the recipe and individual equivalent strategies, including cultural, descriptive and functional, against the baseline results.

Table \ref{tab:equivalent_strategy_comparison} shows that for GPT-3.5, the functional equivalent strategy outperforms others, especially when combined with recipe. For GPT-4o, both descriptive and functional strategies yield better results in CSI-2 and CSI-3, with descriptive strategy excelling in CSI-2 when a recipe is included. In CSI-3, "Recipe + Functional" strategy leads to a significant performance boost of +7.97.

Upon reviewing the translations, both descriptive and functional strategies align well with the gold standards for CSI-2 and CSI-3. However, due to its complexity, the descriptive strategy produces longer translations with trivial details for CSI-3, which is likely to negatively affect COMET scores.

\subsection{Human Evaluation}
\label{sec:evaluation_human}

We collect ratings from 10 native Chinese speakers fluent in English, based on the concept of cross-cultural adaptation on a scale of 0 to 10, alongside automatic quantitative metrics. We select the top-performing methods with recipes, as evaluated by COMET in Tables \ref{tab:evaluation_recipe_adaptaion_main_methods} and \ref{tab:equivalent_strategy_comparison}. We then randomly sample 15 entries with perfect agreement from each CSI category (1: Concrete, 2: Creative, 3: Abstract), totalling 45 entries.

The human evaluation results reveal a trend of performance improvement from CSI-1 to CSI-3 in GPT-3.5 and 4o. We use green to highlight cells with major improvements, i.e. over 1 point. Interestingly, for more complex CSIs (i.e. CSI-2 and CSI-3) we have larger improvements. We also observe that these trends align well with COMET trends in Table \ref{tab:evaluation_recipe_adaptaion_main_methods}, noting that by both metrics, translation theory prompts yield significantly better results than basic prompting across categories.



Interestingly, human evaluators prefer ``\textit{Recipe + Neutralisation}" instead of ``Recipe + Equivalent", the highest in COMET. This preference may stem from the neutralisation definition used in this study, based on the findings of \citet{amenador_wang_2022}. They note that neutralisation is the most common strategy employed by human translators for Chinese names, suggesting a familiarity that could influence the evaluators' preferences towards human-like translation outputs.

\section{Discussion}

The CSI categorisation can be applied to wider cultural domains that contain figurative elements. Future research can use this taxonomy to analyze how different translation methods perform on figurativeness and cultural specificity, suggesting a new framework for evaluating CSI translation. This is similar to the evaluating framework in cultural inference, categorising entailment in different levels to better assess an LLM's ability to understand cultural inference \cite{huang_yang_2023}.

CSI automatic identification offers a cost-effective approach that outperforms GPT-based prompting in CSI-2 and 3. This method is versatile and applicable to both general and domain-specific CSIs, as it focuses on preserving meaning in translation. By doing so, it could enhance the quality of translations in a wide variety of domains where maintaining cultural integrity is essential -- like literature, media, marketing, cross-cultural communication etc.

The findings of this paper also demonstrate the effectiveness of prompt strategies inspired by translation studies in overcoming the challenges of translating CSIs, particularly when direct equivalents are lacking across cultures. This approach shows promise for using LLMs with tailored prompts, integrating human translation insights, and translating diverse cultural elements more effectively.

\section{Conclusion}

In this paper, we introduce the ChineseMenuCSI dataset for CSI-rich dishes and propose a detailed classification in the test set. The results show that LLMs outperformed NMT systems, while NMT is better for Non-CSI translations. Additionally, automatic methods are better than GPT-based prompting at identifying CSIs in most categories.

Incorporating translation studies and recipe details improves LLMs' translation of Chinese dish names. Equivalence strategies, aligned with popular restaurant translations, yield consistently high-quality results, while neutralisation, based on previous analyses, is well-received by evaluators.

\section*{Limitations}

We acknowledge a few limitations of our study. Firstly, we use COMET as the primary automatic evaluation metric for CSIs. While COMET provides a robust evaluation, assessing cultural awareness may require an even deeper understanding of cultural backgrounds in both source and target languages, which COMET may not fully cover. Currently, in the absence of a metric that can evaluate text-to-text cultural similarity, we use COMET due to its high correlations with human judgment. 

Secondly, while we only test zero-shot prompting for translation studies and recipe information, other research, such as \citet{nayak2024benchmarking}, has demonstrated promising results using few-shot in-context learning strategies, which should also be explored.

Lastly, we only sample 45 menu entries from the test set, which can be relatively small compared to the studies with a larger test set. To achieve more robust and reliable results, increasing the number of human evaluators and the sample size of evaluation entries would be beneficial.

\section*{Ethical Considerations}

In conducting this research, we adhere to ethical guidelines to ensure the integrity and responsibility of our work. The ChineseMenuCSI dataset is created by scraping publicly available restaurant websites, ensuring that no private or sensitive information is collected. We obtain data in compliance with the terms of use of the websites and anonymise any identifying details of the restaurants. The human annotators involved in this study are fully informed about the nature of the research and provide their consent. We make the dataset available for research purposes under a license that respects the rights of the original content creators.


\bibliography{custom}
\newpage
\appendix

\section{Appendix}
\label{sec:appendix}

\subsection{Detailed Prompt Teamplates}
\begin{figure}[h]
    \centering
    \begin{tcolorbox}[
        colframe=black!50,
        colback=white,
        sharp corners,
        boxrule=0.5mm,
        width=0.5\textwidth,
        title=\scriptsize\textbf{Prompt Strategy: Recipe + Equivalents}]
        \scriptsize
        \textbf{User:}
        \vspace{2mm}

        \hspace{5mm}Similiar Recipe: [Recipe Instructions].

        \vspace{2mm}
        \hspace{5mm}Based on the above recipe information, provide three translations for [Chinese dish name] based on the three translation strategies listed below and select the best one:
        
        \vspace{2mm}
        \hspace{5mm}Cultural Equivalent: Substituting a source language term with a term from the target language that has similar cultural resonance and functionality.

        \vspace{2mm}
        \hspace{5mm}Functional Equivalent: Rendering the source language's meaning, intent, and style into the target language in a culturally appropriate and understandable way. This strategy prioritizes the effect and function of the text in the target culture over a word-for-word translation, ensuring the translation fulfills the same purpose as the original.

        \vspace{2mm}
        \hspace{5mm}Descriptive Equivalent: Providing an in-depth explanation of a term or concept that lacks a straightforward equivalent in the target language. The explanation could include details such as ingredients, culinary method, key characteristics, etc.

    \end{tcolorbox}
    \caption{Recipe + Equivalents Detailed Prompt}
    \label{fig:prompt_template_equivalent}
\end{figure}

\begin{figure}[h]
    \centering
    \begin{tcolorbox}[
        colframe=black!50,
        colback=white,
        sharp corners,
        boxrule=0.5mm,
        width=0.5\textwidth,
        title=\scriptsize\textbf{Prompt Strategy: Recipe + Neutralisation}]
        \scriptsize
        \textbf{User:}
        \vspace{2mm}

        \hspace{5mm}Similiar Recipe: [Recipe Instructions].

        \vspace{2mm}
        \hspace{5mm}Based on the above recipe information, provide a translation for [Chinese dish name] with the following translation strategy:
        
        \vspace{2mm}
        \hspace{5mm}Menu Description Strategy: This strategy involves using culturally neutral language to describe or explain a cultural word, phrase, or rhetorical expression from the source text (ST). It answers the question, 'What is this?' and is similar to converting a metaphor to its literal meaning. The translations should include the key culinary method, ingredients, and characteristics.

    \end{tcolorbox}
    \caption{Recipe + Neutralisation Detailed Prompt}
    \label{fig:prompt_template_neutralisation}
\end{figure}

\end{document}